\documentclass{article}

\usepackage{arxiv}

\usepackage[utf8]{inputenc} % allow utf-8 input
\usepackage[T1]{fontenc}    % use 8-bit T1 fonts
\usepackage{hyperref}       % hyperlinks
\usepackage{url}            % simple URL typesetting
\usepackage{booktabs}       % professional-quality tables
\usepackage{amsfonts}       % blackboard math symbols
\usepackage{nicefrac}       % compact symbols for 1/2, etc.
\usepackage{microtype}      % microtypography
\usepackage{lipsum}		% Can be removed after putting your text content
\usepackage{graphicx}
\usepackage{cite}
\usepackage{doi}
\usepackage{amssymb}
\usepackage{tabularx}
\usepackage{ltablex}

\title{A Study on Machine Learning Approaches for Player Performance and Match Results Prediction}

\author{Harsh Mittal \\
	Department of Computer Applications\\
	National Institute of Technology\\
	Kurukshetra, Haryana 136119 \\
	\texttt{harsh\_51710010@nitkkr.ac.in} \\
	\AND 
	Deepak Rikhari\\
	Department of Computer Applications\\
	National Institute of Technology\\
	Kurukshetra, Haryana 136119 \\
	\texttt{deepak\_51710013@nitkkr.ac.in} \\
	\AND
	\href{https://orcid.org/0000-0002-2938-6432}{\includegraphics[scale=0.06]{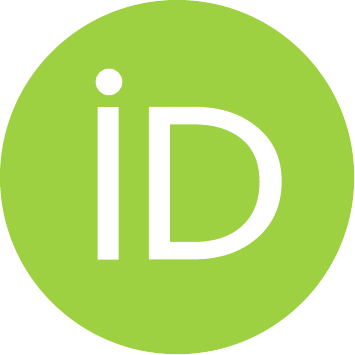}\hspace{1mm}Jitendra Kumar} \\
	Department of Computer Applications\\
	National Institute of Technology\\
	Tiruchirappalli, Tamilnadu 620015 \\
	\texttt{jitendra@nitt.edu} \\
	\AND 
	\href{https://orcid.org/0000-0002-8053-5050}{\includegraphics[scale=0.06]{orcid.pdf}\hspace{1mm}Ashutosh Kumar Singh}\\
	Department of Computer Applications\\
	National Institute of Technology\\
	Kurukshetra, Haryana 136119 \\
	\texttt{ashutosh@nitkkr.ac.in} \\
}

\date{}

\renewcommand{\headeright}{Technical Report}
\renewcommand{\undertitle}{Technical Report}
\renewcommand{\shorttitle}{Player Performance and Match Results Prediction}

\hypersetup{
pdftitle={Machine Learning Approaches for Player Performance and Match Results Prediction},
pdfauthor={Jitendra Kumar},
pdfkeywords={Machine Learning, Prediction, Neural Networks, Support Vector Machine, Cricket},
}

\begin{document}
\maketitle

\begin{abstract}
	Cricket is unarguably one of the most popular sports in the world. Predicting the outcome of a cricket match has become a fundamental problem as we are advancing in the field of machine learning. Multiple researchers have tried to predict the outcome of a cricket match or a tournament, or to predict the performance of players during a match, or to predict the players who should be selected as per their current performance, form, morale, etc. using machine learning and artificial intelligence techniques keeping in mind extensive detailing, features and parameters. We discuss some of these techniques along with a brief comparison among these techniques. 
\end{abstract}

% keywords can be removed

\section{Introduction}
A game of cricket requires two teams, each with 11 players playing consisting of batsmen, bowlers, and all-rounders. A batsman’s main aim is to score as many runs as possible and a bowler’s main aim is to restrict the other team at as least runs as possible by taking more and more wickets. The team which scores more runs at the end of the match is the winner. 

There are many factors which affects the performance of a team and ultimately plays a vital role in deciding whether which team has an upper hand over the other team. These factors include batting first/chasing, time of the match, importance of the match, opposition team, match ground, batting position, current team’s morale, performance statistics, to name a few. Except for the selection of the players for the team, the rest of the factors are generally considered as uncontrollable. So, it is the responsibility of the team’s management, team’s captain, and coach to select the best possible playing XI by considering each player’s current form, performance statistics, fitness, etc.

With the advancement of machine learning techniques and other data analytical tools, researchers have previously tried to predict the outcome of a particular match or even a whole tournament. An ``almost accurate'' prediction would be mostly usable for the cricket playing teams, their management and board of control; to determine a visionary approach to build a team that will be most suitable for their side during world cup and will have most chances of winning.

\section{Techniques}
Machine learning is widely used across the applications domains including detection, regression, classification, identification, optimization etc.~\cite{39,40,41,42,x1,x2,x3,x4,x5,x6,x7,x8,x9,x10,x11,x12,x13,x14}. In this section, a brief discussion on various machine learning approaches is given.

\subsection{k-Nearest Neighbour}
KNN is a supervised machine learning algorithm used for classification problems. It works on the fact that if most of the `k' nearest neighbors of a sample lies in a particular class, then we can assume that the sample also lies in that class~\cite{1}. 

Madan Gopal Jhawar and Vikram Pudi~\cite{2} presented a paper for predicting the outcome of ODI cricket matches using KNN approach. They predicted the performance of each player of both the teams by dividing the players of each team in two groups (batsmen and bowlers), then they used two different algorithms to predict the overall batting and bowling performance and using that result they predict the overall performance of the two teams and used that result to predict the outcome of the match. For batsman’s performance prediction, they used features like matches played, batting innings, and batting average etc. based on these features they calculated the career score and ratio of the number of matches played to the number of matches in which the batsman did batting, this ratio is used to calculate the weight of the performance of that player. After calculating the performance of each batsman, they add them all up to get the batting score of the complete team. The same procedure was followed for bowling to predict the bowler score, then they combined this bowling score with the ratio of the number of matches played to the number of matches in the player bowled to predict the overall performance of a bowler. After getting these batting and bowling performances of two teams they predict the winner.

A. A. Aburas, M Mehtab and Y Mehtab~\cite{3} presented a paper to predict the Cricket World Cup using KNN Intelligent Big Data Approach. They fetched the data from ESPN Cricinfo\footnote{http://www.espncricinfo.com/} and Kaggle\footnote{https://www.kaggle.com/}. After fetching, data was processed by removing the no-entry fields, and non-numeric fields. In data processing, they also used ICC website and expert cricket knowledge to classify the players as either good, bad or elite. They used this data to train their model. After training, they predicted the class of every player in a team. Using the predicted class, they assigned points to each member. Each elite player was given two points and one point was given to each good player. After this assignment, they calculated the sum of all the players for a particular team and associated that sum to the corresponding team. Now the team with the highest score was considered as the strongest and is taken as the predicted winner of the world cup.

\subsection{Logistic Regression}
In logistic regression, the bulk of data is fed to the model. Data has many inputs, where each input tells the value of some features and the class which this input belongs to. Based on the data fed model learns how the values of features are related to the class. Logistic regression model returns a continuous function range of which is [0, 1]. This value is then used to predict the class to which the corresponding input values belong to.

Parag Shah and Mitesh Shah~\cite{6} predicted the outcome of a particular cricket match by considering factors like home field advantage, game plan, match time, whether the match is day-night or day only, etc. Some other factors like average score at the venue, ICC points of the teams, etc. were also considered. They went on to find the probability of winning of a team by assigning weightage to these factors. They used the technique of logistic regression to model the outcome of the match.

\subsection{Neural Networks}
A neural network can be used for classification and pattern recognition problem, all it needs is a large set of data on which it can be trained. The neural ``network'' contains various layers. Every layer consists of various nodes, these nodes are called neurons. Each neuron multiplies the inputs to weights and then adds all the values and pass it to an activation function, the output of which is transferred to the neurons of the next layer. The following layer does the same task with its parameters. This procedure is known as forward propagation. Once we get all the predicted values for all the training data, we compare those values with the actual output and calculate the loss. The main objective of a neural network is to minimize this loss value. This objective is achieved in back propagation.

S. R. Iyer and R. Sharda~\cite{7} tried to predict the performances of the athletes using Neural Networks. They devised an approach to help teams select the best players for the country. They assigned subjective ratings based on heuristic rules to batsmen and bowlers. They called these ratings as ``primary ratings''. Their study showed that neural networks can be used to make a machine learn to classify a bowler or batsman based on his past performances. Muthuswamy and Lam~\cite{8} studied the performance of main Indian bowlers against the seven nations against which the team plays most matches in a calendar year. They used back propagation network and radial basis network function. They mainly predicted that how many wickets a bowler is likely to take and how many runs a bowler is likely to concede in a given ODI match. 

Barath Narayanan~\cite{9} presented a model using an ensemble classification approach and a model using a neural network. He only focused on the historical data of 10 teams that have qualified for World Cup 2019. The article included the 65 features which were used by the model. Some of these features were based on the team while some were based on individual's performance. The classification model used 10 classes, each showing the winning status of one of the teams. In the Neural Network approach, the model used 12 hidden layers. He then trained this model using the data till the 2011 World Cup and then tested it on the data available for the 2015 World Cup. The model predicted that Australia would win by 25.1\% which is highest among all other probabilities, and we know that Australia won the World Cup 2015.

\subsection{Support Vector Machines}
Support Vector Machines, commonly abbreviated as SVM, is a supervised machine learning approach that can be used for both classification and regression purposes. However, they are generally used more for classification problems. In SVM, our main focus is to find a hyperplane that best divides a dataset into two classes. Now to find the best hyperplane, we find the distance between the hyperplane and the nearest data point from either set which is called margin. We choose the hyperplane with the greatest possible margin between the hyperplane and any point within our training set. If we are not able to find a clear segregation, we map our data into higher dimension, say 3D from 2D. Now, instead of a line we will get a plane that divides the data into classes. Similarly, we will keep on mapping data into higher dimensions until we find a hyperplane to segregate it~\cite{10}.

\subsection{Na\"ive Bayes}
Na\"ive Bayes is also used to construct classification and prediction models. Na\"ive Bayes uses the concept of Bayes' theorem. It makes a few assumptions about the data. First assumption is that all the features are independent of each other. The second assumption is that all the features have equal importance. Independent features imply that no feature will affect any other feature. There are cases when features are not generally independent, and some features have more value than other features. In those cases, Na\"ive Bayes should not be used. Na\"ive Bayes predicts quite well when the two conditions of the assumptions are met, and it is faster and easier than other algorithms.

Neeraj Pathak and Hardik Wadhwa~\cite{11} predicted the outcome of a particular ODI match using three machine learning techniques namely Na\"ive Bayesian, Support Vector Machines and Random Forest. They took historical data for the ODI matches between 2001 and 2015 for each time with definite result and successful completion, i.e., matches which were tied or washed away because of rain were ignored. They used Kappa statistic which is used as a performance measure - higher value signifies better performance. They achieved the highest accuracy through SVM. However, the rest of the classifiers were not far enough and achieved a slight less accuracy. Also, these classifiers only performed well when there was no class imbalance. In case of class imbalance, these classifiers failed miserably.

\subsection{Random Forest}
Random Forest is another supervised classification algorithm. It is based on decision tree, which is the basic building block of random forest~\cite{12}.

Sushant Murdeshwar~\cite{13} predicted the outcome of an Indian Premier League (IPL) match by considering the teams’ past performance at a high-level extent instead of considering each player’s individual performance. He basically divided the project into five phases where each phase was considered as a project milestone. The five phases were Data Set Generation, Data Cleaning, Attribute Selection, Data Mining and Analysis of Results. He used attribute selection algorithms like Wrapper method and Ranker method to cut down the number of attributes. Kalpdrum Passi and Niravkumar Pandey~\cite{14} in their paper tried to predict how many runs a batsman will score and how many wickets a bowler will take in a particular match using four different machine learning techniques. They calculated the attributes for each player and assigned weight to them according to its relative importance over other measures using a tool called Analytic Hierarchy Process (AHP). They then calculated the derived attributes using these attributes. These derived attributes consist of attributes such as consistency, form, venue points, etc. After the cleaning of data, it was fed into the four algorithms and the result was analysed. For both datasets namely batsmen and bowlers, Random Forest turned out to be the most accurate classifier. 

Sonu Kumar and Sneha Roy~\cite{15} tried to predict the innings total of a team after 5 overs of the innings in a One-Day International match. They used Linear Regression and MLP Regressor for prediction model and k-Nearest Neighbours, Support Vector Machines, Na\"ive Bayes and MLP Classifier for classification. They argued that the run prediction system used currently by ICC which is based on current rate is flawed and does not predict the total runs accurately. However due to the lack of data at disposal, the machine could not be trained as much as they expected. Tejinder Singh, Vishal Singla, Parteek Bhatia~\cite{16} presented a paper in which they predicted the score and winning team in cricket using Data Mining. They used a dataset that consists of cricket matches played (excluding rain-interrupted matches) between 2002 and 2014 among 8 teams. 

For the team batting in the first innings, they considered all the matches in which that particular team had played first. Furthermore, each match statistics is divided into the 5-over period. From these matches, they considered the features like fall of wickets, run scored, run rate and etc. at each period to predict the runs scored by the first team. For the second team that played in the second innings, they chose only those matches in which that team had played in the second innings and each match was divided into the 5-over period. They also considered the target to be chased and the winning status of the chasing team in terms of `Yes' or `No'.

For the first innings, linear regression has been implemented on the training dataset, with ten-fold cross validation which helps in predicting the score of a team at a particular venue at different situations. For the second innings, Naïve Bayes classifier has been implemented with the same ten-fold cross-validation which gives the probability of winning of the second team at a particular venue at different situations of the match. This probability is used to predict the match outcome.

\subsection{Comparative Analysis}
The following table shows the comparative analysis between different techniques used for the prediction purposes along with the performance metrics devised. 

	\begin{tabularx}{\linewidth}{cXXcXX}
		\multicolumn{6}{c}{Table 1: A detailed comparison fraud detection approaches} \\
		\toprule
		Reference	&	Technique used	&	Data source	&	Pre-processing	&	Performance metric	&	Result	\\
		\midrule
		\cite{2}	&	KNN	&	cricinfo.com	&	$\checkmark$	&	Match result accuracy	&	71\% accurate	\\
		\cite{3}	&	KNN	&	Cricinfo.com, Kaggle.com	&	$\checkmark$	&	Class prediction of Players	&	India and England are top contenders for World Cup, 2019	\\
		\cite{6}	&	Logistic Regression	&	cricinfo.com	&	$\checkmark$	&	Accuracy	&	74.9\% accurate in predicting the match result	\\
		\cite{7}	&	Neural Networks	&	cricinfo.com	&	$\checkmark$	&	Players recommendation accuracy	&	70\% of the players that were recommended were selected for the team	\\
		\cite{8}	&	Backpropagation network and Radial Basis Function Network (RBFN)	&	cricinfo.com	&	$\checkmark$	&	Accuracy of runs scored and wickets taken	&	RBFN performed better with 91.43\% of accuracy	\\
		\cite{9}	&	Neural Network and Ensemble Classification	&	cricinfo.com	&	$\checkmark$	&	Predicted probability of world cup winner	&	Neural network gave better results for 2015 World Cup	\\
		\cite{11}	&	Naïve Bayes, Random Forest and SVM	&	cricinfo.com	&	$\checkmark$	&	Kappa Statistic (Higher Kappa means higher classifier accuracy)	&	SVM performed better than other classifiers	\\
		\cite{13}	&	Random Forest, Naïve Bayes, Decision Trees and KNN	&	cricsheet.org	&	$\checkmark$	&	Match result accuracy using Percentage Split and K-Fold Cross Validation	&	K-Fold Cross Validation performed better with 60-70\% accuracy	\\
		\cite{14}	&	Naïve Bayes, Decision Trees, Random Forest, SVM	&	cricinfo.com	&	$\checkmark$	&	Accuracy in predicting runs scored and wicket taken	&	Random Forest performed better with 90\% test data	\\
		\cite{15}	&	KNN, SVM, Naïve Bayes and MLP Classifier	&	cricinfo.com cricbuzz.com wikipedia.com	&	$\checkmark$	&	Prediction of innings total	&	MLP Classifier performed better	\\
		\cite{16}	&	Linear Regression, Naïve Bayes	&	cricinfo.com	&	$\checkmark$	&	Run rate and Winning percentage prediction	&	For winning prediction, performance of Naïve Bayes improved with increase in overs	\\
		\bottomrule
	\end{tabularx}

\section{Conclusions}
Even after in-depth study of features and keeping in mind all the conditions and factors for a particular match, we cannot say anything about the result of a match for sure as we do not have much data yet for the machine to be trained (as only a few thousand matches are played in total yet). Some nations are relatively newer to start playing international level cricket so there is very less data available for them. But we will try to maximize the accuracy of our prediction using suitable approaches even with limited amount of data. 

Also, a single algorithm can produce different result for different features and different datasets. Each algorithm has its own merits and demerits. We saw that neural network has the highest accuracy rate in most of the cases, but it needs a huge dataset to train the neural network model, so it is considered as an expensive technique. Selection of features and dataset directly affects the performance of the model.


\end{document}